\documentclass[conference]{IEEEtran}
\IEEEoverridecommandlockouts
\usepackage{cite}
\usepackage{graphicx,longtable,subfigure,theorem,amsfonts,amsmath,amssymb}
\usepackage{amsmath,amssymb,amsfonts}
\usepackage{algorithmic}
\usepackage[ruled,linesnumbered]{algorithm2e}
\usepackage{graphicx}
\usepackage{textcomp}
\usepackage{xcolor}
\usepackage{fancyhdr,multirow,caption}
\usepackage{geometry}

\def\BibTeX{{\rm B\kern-.05em{\sc i\kern-.025em b}\kern-.08em
    T\kern-.1667em\lower.7ex\hbox{E}\kern-.125emX}}
\begin{document}

\title{\LARGE On Functional Test Generation for Deep Neural Network IPs}
\author{
    Bo Luo, Yu Li, Lingxiao Wei and Qiang Xu\\
    %CUhk REliable Computing Laboratory (CURE)\\
    Department of Computer Science \& Engineering \\
    The Chinese University of Hong Kong, Shatin, N.T., Hong Kong\\
    %Shatin, N.T., Hong Kong \\
   Email: \{boluo,yuli,lxwei,qxu\}@cse.cuhk.edu.hk
}
\maketitle

\begin{abstract}
Machine learning systems based on deep neural networks (DNNs) produce state-of-the-art results in many applications. Considering the large amount of training data and know-how required to generate the network, it is more practical to use third-party DNN intellectual property (IP) cores for many designs. No doubt to say, it is essential for DNN IP vendors to provide test cases for functional validation without leaking their parameters to IP users. To satisfy this requirement, we propose to effectively generate test cases that activate parameters as many as possible and propagate their perturbations to outputs. Then the functionality of DNN IPs can be validated by only checking their outputs. However, it is difficult considering large numbers of parameters and highly non-linearity of DNNs. In this paper, we tackle this problem by judiciously selecting samples from the DNN training set and applying a gradient-based method to generate new test cases. Experimental results demonstrate the efficacy of our proposed solution.

%As they are becoming increasingly larger and deeper, it is more practical to use a trained neural network intellectual property (IP). However, the functionality of neural networks can be easily damaged by maliciously manipulating their parameters. Therefore, IP vendFors need to provide a way for users to validate its functionality when releasing an IP. In this paper, we propose a novel scheme for validating neural network functionality without providing parameters. To be specific, we validate their functionality by checking the output only. The main challenge is to generate some functional tests for effectively activating parameters so that their alteration can propagate to the output. However it is difficult considering the large amount of parameters and highly non-linearity of deep neural networks. In this work, we obtain functional tests by combining two methods: judiciously selecting from training samples and generating new input with efficient data augmentation. Experimental results demonstrate that our method achieves a good tradeoff between detection coverage and cost.

\end{abstract}

\section{Introduction}\label{sec:introduction}

Artificial intelligence (AI) systems based on deep neural networks (DNNs) have achieved great success in many areas such as computer vision, speech recognition and natural language processing. Over the years, neural networks become increasingly larger and deeper, which requires significant amount of data and time to train.
For example, it may take weeks to train a state-of-the-art model with the latest GPUs on the ImageNet dataset~\cite{deng2009imagenet}.
Consequently, it is more practical for individual users or small firms to use a trained DNN intellectual property (IP) (e.g., face recognition module) that is commercially available. In most cases, vendors would prefer a blackbox IP model to protect the architecture and the trained parameters of the DNN.

%DNNs, however, are susceptible to various kinds of attacks. Adversarial example attack~\cite{goodfellow2014explaining,papernot2016limitations,luo2018towards} targets to change the output of neural networks by slightly perturbing the input which is hardly detected by human eyes. And data poisoning attack~\cite{yang2017generative,munoz2017towards,biggio2012poisoning} degrades the performance of neural networks by adding malicious samples into the training set. Recently, Liu \emph{et al.}~\cite{liu2017fault} proposes to directly manipulate the parameters of neural networks to achieve malicious objectives. These attacks are serious threats to safety-critical applications, such as disease diagnosis systems and self-driving systems~\cite{venkatalakshmi2014heart,bojarski2016end}. Therefore, IP vendors need to provide test cases for IP users to validate functional correctness and integrity.

DNNs, however, are susceptible to various kinds of attacks. Adversarial example attacks~\cite{goodfellow2014explaining,papernot2016limitations,luo2018towards} target to change the outputs of DNNs by slightly perturbing their inputs. Recently, there is an increasing number of attacks that target at DNNs themselves instead of their input data. Liu \emph{et al}.~\cite{liu2017fault} first proposes to attack DNN parameters for misclassifications based on two fault injection methods: single bias attack and gradient descent attack.
%\cite{breier2018practical} performs practical laser fault injection to flip bits of activation functions in DNNs.
Reverse-engineering attacks~\cite{hua2018reverse,wei2018know} on hardware DNN accelerators can identify the model parameters in the off-chip memory and then attackers may stealthily substitute original parameters with malicious ones. These attacks seriously threat safety-critical applications based on DNNs. Therefore, it is essential for IP users to validate the functionality of DNNs before everyday usage.

%or the local adversary process orthe local adversary in the IP user's end, and the IP vendor is considered to be trusted.

%Therefore, to ensure the model authenticity, it is essential for IP vendors to provide test cases for functional validation to end users without showing them model details (e.g., architectures and parameters). such as disease diagnosis systems and self-driving systems~\cite{venkatalakshmi2014heart,bojarski2016end}.

Traditional integrity checking methods~\cite{ohkubo2003cryptographic,venkatesan2000robust} based on generating signatures are not applicable for DNN IPs, because IP users can not directly get the model parameters for signature generation. Hardware testing techniques for troubleshooting design defects~\cite{DBLP:journals/corr/PeiCYJ17,ma2018combinatorial} are not applicable either, as IP users have no access to intermediate results of DNNs. To tackle the above problem, in this work, we propose a practical validation scheme for IP users considering their limited black-box access. The idea is for IP vendors to generate functional tests to activate parameters in the DNN whose perturbations will propagate to the outputs. Then, malicious perturbations of model parameters can be directly detected by IP users, just checking the outputs of the functional tests.

%malicious parameters perturbations will not propagate to the output for normal inputs, but only for the specified input designed by the adversary.
However, DNNs are highly generalized models and use non-linear activation functions, only partial parameters will be activated and take effect in the calculation for an input sample~\cite{glorot2011deep}, thus one functional test can only validate part of parameters. Considering the large number of parameters in today's DNNs, it is challenging to generate a reasonable size of functional tests to achieve a high validation coverage.
%activate large numbers of parameters so that their perturbations can propagate to the outputs and then be detected.
In this paper, we solve this problem with two techniques: first, we judiciously select test cases from the existing training set, and when this method becomes inefficient, a novel gradient-based technique is presented to generate new test cases. Experimental results show that the proposed functional test generation method is effective and efficient, achieving a high validation coverage with limited test cases, under both malicious and random perturbations
of DNN parameters. %and the detection rates for both malicious and random parameter perturbations are high.

To the best of our knowledge, this is the first work for functional validation of DNN IPs considering end users black-box access.
%without showing them model parameters.
The main contributions of this work include:
\begin{itemize}
\item We formulate the functional validation of DNN IPs as an optimization problem, wherein we try to generate a small number of test cases that can activate as many parameters as possible.
\item We propose to judiciously select functional tests from the training set in an iterative manner to efficiently activate DNN parameters.
\item We present to generate new functional tests with a novel gradient-based method when selecting from the training set is inefficient.
    %activate the remaining parameters whenever possible.% after the saturating of selecting from training samples.
%\item To the best of our knowledge, this is the first functional test generation work for deep neural network IPs. We formulate the problem as an optimization problem, wherein we try to find a small set of tests that activate as many neuron parameters as possible and propagate to the outputs. %We define the detection coverage of an input pattern, and formulate our functional authentication problem to an optimization problem, finding small number of  inputs to achieve a high detection coverage in neural networks.
%\item Considering the highly generalized feature and non-linear function of neural network, we analyze the characteristics of fault propagate in neural networks. Then we formalize our integrity checking problem to find an input pattern which can activate as many neurons as possible and detect the perturbations with high confidence.
%\item We propose an effective and efficient solution to solve the above optimization problem, where we first select training samples to activate neural networks in an iterative manner, and when the method saturates, we use a novel data augmentation method to generate new tests to activate the remaining parameters whenever possible. %Our experimental results show the efficacy of our scheme.
\end{itemize}

The rest of the paper is organized as follows. In Section~\ref{sec:preliminary}, we give a preliminary introduction about neural networks and the related work. Then we give an overview of our functional test generation scheme in Section~\ref{sec:overview}. Next, the proposed efficient functional test generation method is introduced in Section~\ref{sec:solution}. Finally, we present the experimental results and conclude our work in Section~\ref{sec:experiment} and Section~\ref{sec:conclusion}, respectively.

\section{Preliminaries}\label{sec:preliminary}
\subsection{Neural Networks}

%\begin{figure*}[!ht]
%    \centering
%    \includegraphics[scale=0.66]{./fig/nn_1.pdf}
%
%\caption{(a) A 4-layer deep neural network. (b) The general structure of a neuron. (c) ReLU and Tanh activation functions.}
%\label{fig:neural_network}
%\end{figure*}

Neural networks are organized as successive layers of neurons which are connected by links with different parameters. Each neuron in the hidden layer applies a non-linear activation function on the weighted sum of its input. The output of layer $l+1$ is denoted as:
\begin{equation}
o^{(l+1)} = \sigma(W^{(l)}o^{(l)}+b^{(l)}),
\end{equation}
where $\sigma$ is the activation function. $o^{(l)}$, $W^{(l)}$ and $b^{(l)}$ are the outputs, weights and bias of the $l$-{th} layer, respectively. Weights and bias are called parameters of the network. In this way, the outputs of the current layer are computed by a non-linear function applied on the outputs of the previous layer and its parameters. Usually, there are many layers in DNNs to achieve high generality. Therefore, DNN as a whole is a complex non-linear function of parameters and the input.

Activation functions provide non-linearity so that neural networks can approximate arbitrary functions. There are several activation functions in modern neural networks, such as ReLU and Tanh, which both have some regions of saturation or inactivation~\cite{han1995influence,nair2010rectified}. For example, the output of ReLU will always be zero as long as its input is negative. As neural networks are trained to fit the large training set where the training samples vary a lot to each other, an input sample can only activate partial parameters in a well trained model~\cite{glorot2011deep}.

%The training process of neural networks is guided by a loss function, which calculates the gap between the ground truth and the predicted value. In each training iteration, it updates the parameters according to the gradients of the loss function to decrease the loss and increase the prediction accuracy, until the model gets the expected accuracy~\cite{shen2005loss}. There are several popular loss functions used in machine learning theory, like cross entropy loss, log softmax loss and so on.
\subsection{Related Work and Motivation}
Past work has introduced several ways to inject faults into DNNs themselves for compromising their functionality. In~\cite{liu2017fault}, attackers fool DNNs to make mistakes by modifying their parameters through fault injection, in which single bias attack modifies one parameter with a large perturbation for misclassification and gradient descent attack considers stealthiness by adding small perturbations on a number of parameters. Reverse-engineering attacks~\cite{hua2018reverse} can identify model parameters in the off-chip memory, which may be stealthily replaced by attackers.  \cite{breier2018practical} performs practical laser fault injection on activation functions of DNNs using a near-infrared diode plus laser.

To the best of our knowledge, there exists few work defending against the above functionality compromised attacks for DNN IPs. Traditional signature-based integrity checking methods~\cite{ohkubo2003cryptographic,venkatesan2000robust} are not applicable as IP users can not access the DNN parameters. Testing techniques~\cite{DBLP:journals/corr/PeiCYJ17,sun2018testing,gehr2018ai,ma2018combinatorial} generate test cases to cover all neurons so that design defects of hardware DNNs can be detected and located. However, they are not appropriate for functional validation of DNN IPs under attacks for two reasons: first, IP users have no access to the intermediate model results as system testers do. Second, testing only considers the neuron coverage, which is not enough for covering model parameters under malicious attacks. For example, there are two neurons in adjacent layers that are covered by two separate test cases and no other tests cover them during the testing process. Even though the two neurons can be tested, the attacks targeting to perturb the weight between them cannot be detected. As the two neurons are never activated at the same time with test cases, the malicious perturbations on the weight will never be revealed, but it may cause misclassifications for other inputs.

%As DNNs are widely adopted in safety-critical applications, ensuring their functionality and correctness has received many research efforts recently~\cite{DBLP:journals/corr/PeiCYJ17,sun2018testing,gehr2018ai,ma2018combinatorial}. DeepXplore~\cite{DBLP:journals/corr/PeiCYJ17} first presents a white-box framework to systematically generate test cases that cover most neurons in the network. DeepCover~\cite{sun2018testing} proposes four test criteria inspired by the MC/DC coverage criterion, and then they generate test cases for each criterion based on linear programming. However, DeepCover is only applicable to small-scale neural networks. Gehr \emph{et al}.~\cite{gehr2018ai} proposes the verification approach on DNNs based on abstract interpretation, design specific abstract domains and transformation operators. Most recently,~\cite{ma2018combinatorial} adopts the combinatorial testing techniques on DNNs and generate test cases based on the proposed coverage criteria.

Motivated by the above, in this paper, we propose to validate the functionality of DNN IPs by effectively generating a small number of test cases that can activate model parameters whenever possible and propagate their perturbations to the outputs. IP users only have to run these test cases and check the final outputs of DNNs to validate their functionality without knowing model details. To the best of our knowledge, this is the first work of functional validation for DNN IPs under malicious attacks targeting at model parameters, as detailed in the following sections.

\section{DNN IPs Validation Methodology}\label{sec:overview}

As discussed in previous sections, IP users can just use the DNN IP as a black box: feed the IP with an input and get the corresponding output. Based on this, we propose a practical functional validation scheme for IP users, in which IP vendors will first generate a small number of functional tests and share them with IP users, then users validate the functionality of the IP by checking whether it functions correctly with the shared tests.

\begin{figure}[!h]
    \begin{flushleft}
    \includegraphics[width=1.02\columnwidth]{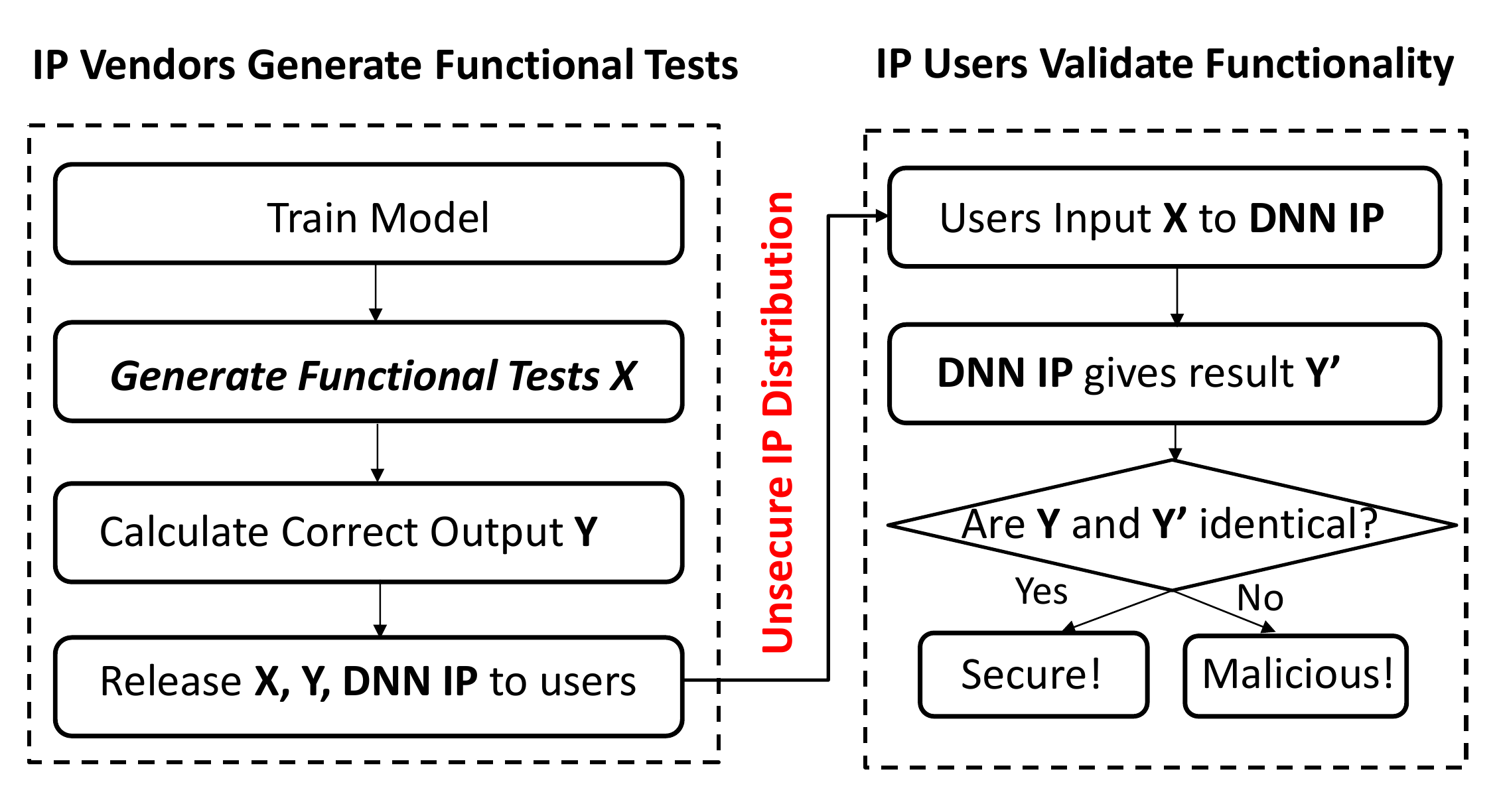}
\caption {The overview of functional validation for DNN IPs.}\vspace{-10pt}
\label{fig:overview}
\end{flushleft}
\end{figure}

The working flow of the proposed functional validation scheme is shown in Fig.~\ref{fig:overview}, which consists of two phases. Firstly, for the IP vendors, they generate a small set of functional tests $X$, then release these test cases, the corresponding outputs $Y$ together with the IP to users. After through an unsecure distribution process, IP users receive the IP and run it as a black box with these functional tests $X$. Then they compare the current outputs $Y'$ with the provided ones $Y$. If they are not identical, the DNN IP has been perturbed. Otherwise, it is secure. The shared functional tests $X$ and the corresponding outputs $Y$ are encrypted, thus their integrity can be ensured.

As DNNs are extremely complex and non-linear, only partial parameters take effect for one test case. The perturbations of other parameters will not be detected and thus not be validated with this test. Therefore, the key challenge in our validation scheme is to effectively generate a reasonable number of functional tests that can validate DNN parameters as many as possible under malicious perturbations, which is demonstrated in the next section.

%As DNNs are extremely complex and non-linear, perturbations of some parameters will not affect the outputs of normal samples, but just cause malicious result for specific input designed by the adversary. So using normal samples to do functional validation can not detect these kinds of perturbations.

%\textbf{Thus, the technical challenge in our validation scheme is for IP vendors to generate functional tests which can fully validate the DNN IPs under both malicious and random perturbations.} We propose an efficient and effective solution for this problem which is the main contribution, detailed in the next section.

\section{Efficient Functional Test Generation}\label{sec:solution}
%In this section, we present the functional tests generation method for IP vendors which can fully validate the malicious and random perturbations of DNN IPs. Firstly, we define the validation coverage metric which identifies the validation capability of functional tests. Then, we present the functional tests generation method based on the validation coverage metric.

In this section, we first define our validation objective. Then we propose to judiciously select tests from the existing training set and when this method becomes inefficient, a new gradient-based test generation technique is presented. Finally, these two approaches are combined in a unified way to efficiently generate functional tests for DNN IPs.

\subsection{Validation Objective}\label{sec:vc}
In our validation scheme, we call a parameter is activated when perturbations of it will propagate to the DNN output and be detected. Otherwise, it is un-activated. A parameter can be validated when at least one test case activates it. As the gradient of a function measures the sensitivity of its output with respect to the change of its argument, we use the gradient of the DNN output with respect to the parameter to determine whether the parameter is activated or not. Assume ReLU is the activation function, given an input $\boldsymbol x$, we define the parameter $\theta_i$ is activated if it satisfies:

%As introduced in previous sections, we know that an input pattern can only activate part parameters whose perturbations will propagate to the DNN output. For the un-activated parameters, their perturbations will not influence the DNN output thus their integrity can not be validated. We propose a math formulation to identify whether a parameter can be validated or not by an input pattern in our validation scheme, it formulates as follows:
\begin{equation}
 \nabla_{\theta_i} F(\boldsymbol x) \ne 0,
\end{equation}
where $F$ is the function of the DNN. $\nabla_{\theta_i} F(\boldsymbol x)$ calculates the gradient of $F(\boldsymbol x)$ with respect to $\theta_i$. %when $\nabla_{\theta_i} F(x)$ does not equal to zero, it means that the parameter is activated.
Unlike ReLU, the gradients of other activations (e.g., Sigmoid and Tanh) in the saturation regions are quite small and approximate to zero. In this case, we define $\theta_i$ is activated when $\nabla_{\theta_i} F(\boldsymbol x)$ is greater than a small value $\epsilon$. For the easy explanation of our method, we assume ReLU is the default activation function.

Therefore, the validation coverage of a functional test $\boldsymbol x$ can be formulated as follows:
\begin{equation}
VC(\boldsymbol x)=\frac{\#\{\theta \; | \; \nabla_\theta F(\boldsymbol x) \neq0\}}{\#(\theta)},
\end{equation}
where the numerator is the number of activated parameters, and the denominator is the number of total parameters in the DNN. The validation coverage of a functional test equals to the percentage of parameters it activates.

As one functional test can only activate partial parameters, it is necessary to use a set of functional tests to achieve a high validation coverage. Given a test set $X$ with $n$ samples, its validation coverage is as follows:
\begin{equation}
VC(X) = \frac{\#(P_1 \cup P_2 \cup ... P_n)}{\#(\theta)},
\end{equation}
where
\begin{equation}
P_i =\{\theta \; | \; \nabla_\theta F(\boldsymbol x_i) \neq 0\}.
\end{equation}

$P_i$ denotes the parameter set activated by the test case $\boldsymbol x_i$, and the validation coverage of $X$ is the percentage of unique parameters activated by all tests in $X$.

%\subsection{Validation Goal}\label{sec:vc}
%Considering the large number of parameters and the high non-linearity of DNNs, to achieve a high validation coverage requires a large number of functional tests and induce a high validation cost. Thus, the functional tests generation method should make a good trade off between the validation coverage and the validation cost, we formulate the tradeoff as follows:
Generally speaking, more test cases can activate more parameters and thus obtain a higher validation coverage, but will incur a larger validation cost. Therefore, it is essential to achieve a good tradeoff between the validation coverage and cost. We formulate this problem as follows:
\begin{equation}
     \begin{gathered}
        \mathop{\arg\max}_{X}\  VC(X)\\
        s.t. \quad \#(X) \le N_t,\\
     \end{gathered}
\end{equation}
where $N_t$ is the maximum number of test cases allowed for functional validation. Our objective is to maximize the validation coverage with a limited number of test cases. Next, we introduce techniques to solve this problem in detail. % The goal of the tests generation method is to generate a functional set $X$ which maximizes the validation coverage with a limited number of functional tests.
%However, considering the large number of parameters and highly non-linearity in todays' DNN, solving this problem is challenging.
%To find the efficient functional set, making a good trade off between the detection coverage and the cost, we propose two heuristic methods, illustrated in the next section.
%In our method, we achieve this objective by efficiently generating a reasonable number of functional tests that activate as many parameters as possible. We detail our solution in the next section.

\subsection{Selecting from Training Set}

The first solution we propose is to select functional tests from the existing training set based on the following heuristic: as the DNN is trained to successfully perform some tasks (e.g., regression and classification) on the training set, most parameters will participate in processing these tasks. In other words, if many parameters are not activated in the training set, the network is not trained well, as many resources are wasted. %This heuristic is evaluated correct in the following experimental results in Section 5.

Based on the above analysis, we judiciously select test cases from the training samples in an iterative manner. In each iteration, we choose the sample that can activate the maximal number of un-activated parameters. At the beginning, the chosen validation set is empty, and the sample with the highest validation coverage is firstly selected. Then in the following iterations, we choose the next sample $\boldsymbol s_i$ from the training set $S$ according to the following equation:
\begin{equation}
\mathop{\arg\max}_{\boldsymbol s_i\in S}\ VC(X+ \boldsymbol s_i)- VC(X),\\
\end{equation}
where $X$ is the current validation set that includes the chosen samples in previous iterations. This equation selects the input that can activate the most un-activated parameters or lead to the largest validation coverage increase.

\begin{algorithm}[h]
{\fontsize{9pt}{9pt}\selectfont
\KwIn{DNN function $F$, training set $S$, maximum functional tests $N_t$.}

\KwOut{Validation set $X$.}
Initialize validation set: $X=\varnothing$;\\
\While{$\#(X) < N_t$}{
    \For{$\boldsymbol s_i \in S$}{
        $\Delta_{\boldsymbol s_i}VC=VC(X+\boldsymbol s_i)-VC(X)$;}

    	Select $\boldsymbol s_i$ with the largest $\Delta_{\boldsymbol s_i}VC$;\\
    	Add $\boldsymbol s_i$ to the validation set $X$;\\
        Update $VC(X)$.\\
}
\caption{Selecting from training set.}\label{alg:train}
}
\end{algorithm}

The whole process of selecting functional tests from the training set is shown in Algorithm~\ref{alg:train}, in which we first initialize the validation set as empty. During each iteration, we calculate the benefit or the increased validation coverage $\Delta_{\boldsymbol s_i}VC$ achieved for each training sample in line 3-5. Then we select the best one which brings the largest validation coverage increase, and add it to the validation set $X$ in line 7. The iteration is continued until the number of functional tests exceeds the limit $N_t$.

%
%The experimental results in Section 5 found that the method is effective at early iterations, only 20 selected training samples can activate about 91\% parameters for the MNIST model. However, in late iterations, the detection coverage will increase extremely slow, as adding 10000 more samples can only activate additional 4\% parameters. That is to say, the method will saturate quickly.

The experimental results in Section~\ref{sec:experiment} show that this method is effective at early iterations, achieving a high validation coverage with a very small number of functional tests. However, in late iterations, the validation coverage will increase extremely slow with new functional tests added. That is to say, the method will saturate quickly.
%The possible explanation is as follows: the neural network is trained for increasing the generalization capability on unseen data that comes from the same unknown distribution as the training data. Some features of the data from the unknown distribution may not frequently exist in the training data. In other words, some neurons or parameters in the network may be reserved for generalization and they can only be activated with very rare features that even do not exist in the training data.
To solve this problem, next we propose to generate new samples to activate the remaining parameters as many as possible when training samples are no longer efficient.

\subsection{Gradient-based Test Generation}
Considering there are some parameters difficult to activate by the training samples, we propose to generate new samples to activate these bottleneck parameters. The key idea is to generate \emph{synthetic training samples} which can be classified correctly by the network consists of the un-activated parameters. The intuition is that samples correctly classified by a DNN will have similar features with its training samples, thus can efficiently activate the network parameters. Based on this, we propose to efficiently activate the bottleneck parameters by generating synthetic training samples based on the gradient descent technique widely used for training DNNs.

Unlike training DNNs, wherein parameters are updated to minimize the loss, we update the input to reduce the loss according to the gradients of it. This can be formulated as follows:

%As the loss function in DNN calculates the gap between the model prediction and the ground truth, thus we can generate the synthetic training sample by decreasing the prediction loss. We use the gradient descent method to quickly generate the synthetic samples, updating the input according to the gradients of loss function.
%The gradient descent based tests generation is in the following equation:
\begin{equation}
 \boldsymbol x_i^* = \boldsymbol x_i - \eta \nabla_{\boldsymbol x_i} J(\boldsymbol x_i, y_i, \theta),
\end{equation}
where $J(\boldsymbol x_i, y_i, \theta)$ is the loss function that measures the gap between the model output for an input $\boldsymbol x_i$ and the corresponding ground truth $y_i$. In each update, we change the input with the step size $\eta$ at the directions based on the gradients of $J(\boldsymbol x_i, y_i, \theta)$ with respect to $\boldsymbol x_i$, in which the loss can decrease most quickly. After several iterations, we can get the synthetic training samples that can be classified correctly by the network with un-activated parameters.

%The data augmentation method is inspired by the discovery in Section 4.2, that a small number of training samples can activate most parameters. We explain that a small number of training samples can contain most features of the whole training data, thus they will activate most parameters and neurons in the trained model. Based on this discovery, we propose a simple yet effective data augmentation method to activate the parameters which are difficult to activate by the training set. The key idea is to generate an input set which can be highly classified by the remaining network with the un-activated parameters. In this way, the generated input set will contain the most features for the model to classify, thus largely activate the parameters. We can continue this process until the target coverage is satisfied.

%Recall the learning process, the parameters are updated based on the gradients of loss function to reduce the prediction error. In our case, the parameters are fixed, and we update the input to reduce the classification error. Similarly, we calculate the gradients of the loss function with respect to the input, which indicates how the loss will change when perturbing the input.

In each iteration, we generate a batch of $k$ synthetic training samples where $k$ is the number of the neurons in the output layer. We do this because for classification, the number of neurons in the output layer corresponds to the number of classification categories. Each category has their own unique features and a batch of input containing all these categories will have a higher probability to activate more parameters. %This process can be repeated to generate many batches of input patterns and continually increase the validation coverage.

\begin{algorithm}[h]
{\fontsize{9pt}{9pt}\selectfont
\KwIn{Loss $J$, category number $k$, maximum functional tests $N_t$, maximum gradient descent updates $T$.}

\KwOut{Validation set $X$.}
Initialize validation set: $X=\varnothing$.\\ %($x_i^*$ is the generated sample highly classified to category $i$ in a remaining network).}
\While{$\#(X) < N_t$}{
Initialize $\boldsymbol x_1^*$, $\boldsymbol x_2^*$, ... , $\boldsymbol x_k^*$ with all zeros;\\
$t=0$; \\
\While{$t < T$}{
\For{$i \gets 1$ \KwTo $k$}{
   $\delta=\eta \nabla_{\boldsymbol x_i^*} J(\boldsymbol x_i^*,y_i,\theta)$;\\
   $\boldsymbol x_i^*=\boldsymbol x_i^*-\delta$;\\
   }
$t=t+1$;\\
}
Add $\boldsymbol x_1^*$, $\boldsymbol x_2^*$, ... , $\boldsymbol x_k^*$ to $X$. \\
}

\caption{Gradient-based test generation.}\label{alg:process}
}
\end{algorithm}

The overall process of gradient-based test generation is shown in Algorithm~\ref{alg:process}, where in each iteration, we generate $k$ input patterns, classified as $k$ different categories, respectively. First, in line 3, the inputs are initialized with all zeros. Then, we update these inputs using gradient descent method to iteratively decrease the loss function $J$ in line 5-11. After $T$ iterations, the generated $k$ tests can be classified by the model correctly and we add them to the validation set $X$ in line 12. The process is continued until the number of generated functional tests reaches to the limit.

\subsection{Combined Functional Test Generation}
As Algorithm~\ref{alg:train} is effective at early iterations but quickly becomes inefficient, Algorithm~\ref{alg:process} can continually increase the validation coverage, but is not as efficient as Algorithm~\ref{alg:train} in the early stage (the true training samples are more effective than the synthetic ones). Therefore, we propose to combine these two functional test generation techniques in a unified way, where we generate tests with Algorithm~\ref{alg:train} first, and then switch to Algorithm~\ref{alg:process} when Algorithm~\ref{alg:train} is inefficient.
However, the remained problem is to identify the switch point. We propose to compare the benefit achieved by each algorithm. When the increased validation coverage per test case generated by Algorithm~\ref{alg:process} is greater than the one generated by Algorithm~\ref{alg:train}, we will transform to gradient-based test generation method.
%
%
%The increase of the detection coverage achieved by the $i^{th}$ selected training sample in algorithm 1 is denotes as:
%
%\begin{equation}
%IN_1=max(TC(X+s_i)-TC(X)),
%\end{equation}
%where $s_i$ is the training sample in the training set.
%
%As we generate $n$ samples for all categories in Algorithm 2, we compute the benefit of one generated sample as the average detection coverage:
%\begin{equation}
%IN_2=\frac{\sum_{i=1}^{n} TC(x_i^*)}{n},
%\end{equation}
%where the $x_i^*$ is the generated sample for category $i$.
%When $IN_2 > IN_1$, we switch to Algorithm 2 for generating functional tests.

\section{Experimental Results}\label{sec:experiment}
%We evaluate the proposed validation scheme from two aspects. First, we evaluate the validation coverage of the generated functional tests. Then we evaluate the detection rate of the validtion scheme under both malicious and random parameters perturbations.

%\textbf{Setup:}
\subsection{Experimental Setup}
The experiments are performed with MNIST~\cite{lecun2010mnist} and CIFAR-10~\cite{krizhevsky2014cifar} datasets. The MNIST includes 70000  hand-written digit images, and the CIFAR-10 contains 60000 color images of natural objects. To verify that our validation scheme can apply to varying DNN architectures and activation functions, we train the MNIST model with Tanh activation function, and the CIFAR-10 model with ReLU.

For each dataset, we implement one DNN model, detailed in Table~\ref{tab:model}. The MNIST and CIFAR-10 models achieve 98.9\% and 84.26\% classification accuracy respectively, which are comparable to the state-of-the-art results.
%It should be noted that for Tanh activation function, the gradient of Tanh will be very small when it saturates, not equal to 0 as ReLU. Thus the equation to calculate the validation coverage is set as follows:
%\begin{equation}
%VC(x)=\frac{\#\{\theta \; | \; \nabla_\theta F(x) > 10^{-6}\}}{\#(\theta)}.
%\end{equation}

\begin{table}[!h]
\centering
\begin{center}
\scalebox{1}[1]{
\begin{tabular}{|p{1cm}| p{3cm}| p{3cm}|}
 \hline

  Layer & MNIST &CIFAR \\
  \hline
  1&28*28 Image&32*32 RGB Image\\
  \hline
  2& Conv(3,3,32). Tanh&Conv(3,3,64). ReLU \\
  \hline
  \multirow{2}{*}{3}& Conv(3,3,32). Tanh&Conv(3,3,64). ReLU \\
  &Max pooling(2,2)&Max pooling(2,2)\\
  \hline
  4 & Conv(3,3,64). Tanh&Conv(3,3,128). ReLU\\
  \hline
  \multirow{2}{*}{5} & Conv(3,3,64). Tanh&Conv(3,3,128). ReLU\\
  &Max pooling(2,2)&Max pooling(2,2)\\
  \hline
  6&Fully connect 128. Tanh & Fully connect 512. ReLU\\
  \hline
  7&Fully connect 10.  & Fully connect 10. \\
  \hline
  \multicolumn{3}{|c|}{Softmax}\\
 \hline
\end{tabular}
}
\end{center}
\caption {The architectures of the two models.}\label{tab:model}\vspace{-13pt}
\end{table}

\subsection{Validation Coverage}
In this section, we evaluate the validation coverage of the proposed functional test generation method. %effectiveness of the generated functional tests considering their validation coverage.
%We first show the performance of images from difference sources, and then discuss the detection coverage of our combined activation scheme, comparing with the two component methods.

\subsubsection{Validation Coverage of Different Image Sets}
\begin{figure}[!h]
    \centering
    \includegraphics[width=0.72\columnwidth]{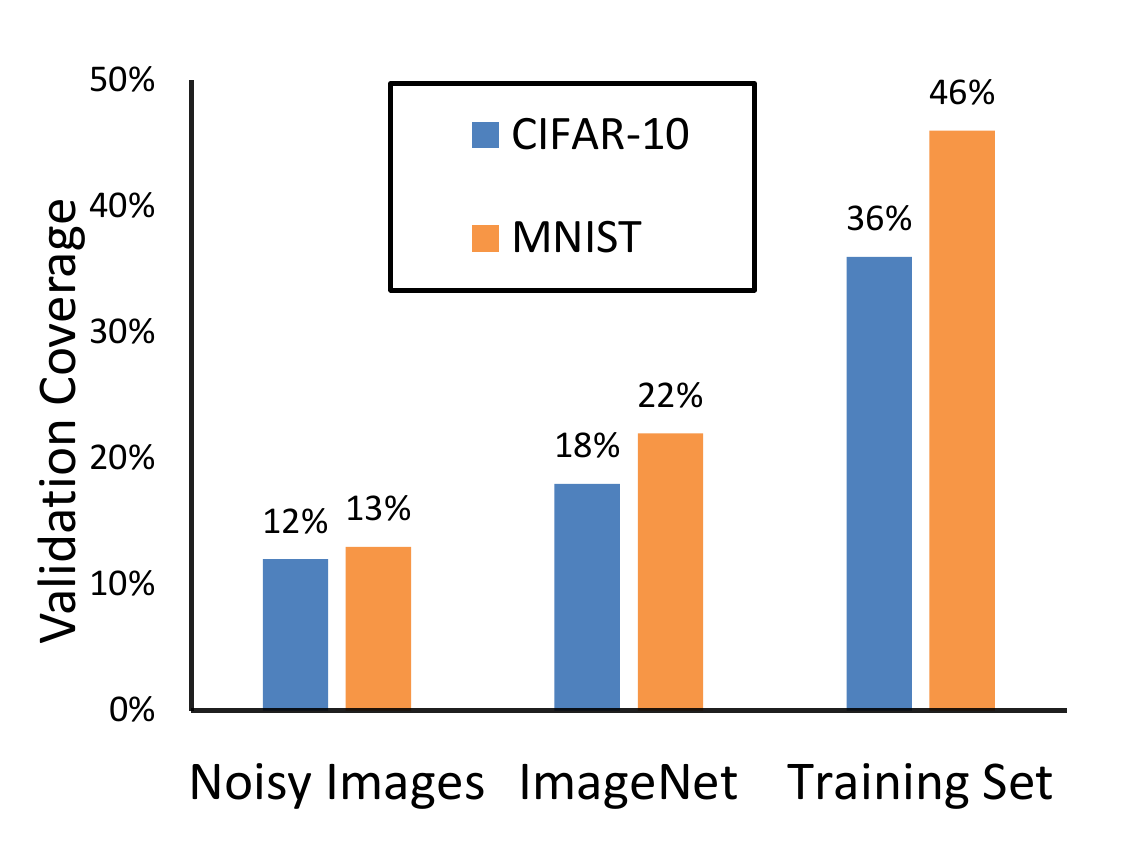}
%\captionsetup{labelfont=bf,textfont=bf}
\caption[textfont]{Validation coverage of different image sets.}\vspace{-5pt} %Firstly, neural vendor will generate test inputs and correct outputs, sharing with IP users. Then IP users do functionality validation by comparing Neural IP results with the shared outputs.}
%The $x$ scale is the number of training samples selected, the $y$ scale is the total detection coverage of the selected samples.}
\label{fig:test_coverage}
\end{figure}

Fig.~\ref{fig:test_coverage} shows the validation coverage of three different image sets: the first one is the noisy images of Gaussian distribution; the second is the ImageNet that is the largest data set in the image recognition area~\cite{deng2009imagenet}; the third is the training set of the corresponding model. For each image set, we randomly select 1000 images and calculate their average validation coverage.

%To illustrate that training samples can efficiently activate the DNN parameters compared with other image sets, we do following experiments: we calculate the validation coverage of different image sets, one is the noisy images of gaussian distribution, the second is the ImageNet set which is the largest data set in the image recognition area, and the third is the training set for the corresponding model. For each image set, we randomly select 1000 images and calculate their average validation coverage.
%The results for CIFAR-10 model are in Table 1.
%The results are shown in Figure~\ref{fig:test_coverage}.

%\begin{figure}
%  \begin{minipage}[t]{0.5\linewidth}
%    \centering
%    \includegraphics[width=2.2in]{./fig/figure1.pdf}
%    \caption{Validation coverage of different image sets for MNIST and CIFAR-10 models.}
%    \label{fig:side:a}
%  \end{minipage}%
%  \begin{minipage}[t]{0.5\linewidth}
%    \centering
%    \includegraphics[width=2.2in]{./fig/figure2.pdf}
%    \label{fig:side:b}
%  \end{minipage}
%\end{figure}

We can see that the training samples achieve the highest validation coverage for both the MNIST and CIFAR-10 model with 46\% and 36\%, respectively. And the ImageNet achieves the second best performance, while random images achieve the worst, where the validation coverage is only 13\% for the MNIST and 12\% for the CIFAR-10.
The results correspond to our analysis that DNNs will take full advantage of their resources (e.g., parameters) to finish the classification task on training samples. As a result, images from the training set will have a higher probability to activate more parameters than others. % will activate most parameters and achieve high validation coverage. %st parameters will be activated to perform the classification task  have more features which can be recognized by the trained model thus they will activate more parameters in the classification process.
Noisy images have little features similar to the training samples and thus activate the least number of parameters.% their randomness.
%thus they are rarely recognized by the model.
%it is not necessary for neural networks to try their best to find useful patterns in noisy images for classification.% it is that they have little feature can be recognized by the model for their randomness.

\subsubsection{Validation Coverage of Different Methods}
Fig.~\ref{fig:test_coverage2} shows the validation coverage of the proposed three functional test generation methods for the CIFAR-10 model, in which we can see that a small number of selected training samples can achieve a high validation coverage, for example, only 20 functional tests can obtain up to 82\% validation coverage. However, selecting from training samples will become inefficient quickly. The validation coverage only increases 4\% when the number of functional tests increases from 20 to 10000. Moreover, we find that there are about 8\% of parameters always un-activated when using the whole training set. We analyze this phenomenon that DNNs are highly generalized models and some parameters are reserved for samples unseen in the training set.
%We evaluate the validation coverage of the functional tests generated by our three proposed methods, which are shown in Figure~\ref{fig:test_coverage2} for the CIFAR-10 model.
\begin{figure}[!h]
    \centering
    \includegraphics[width=0.72\columnwidth]{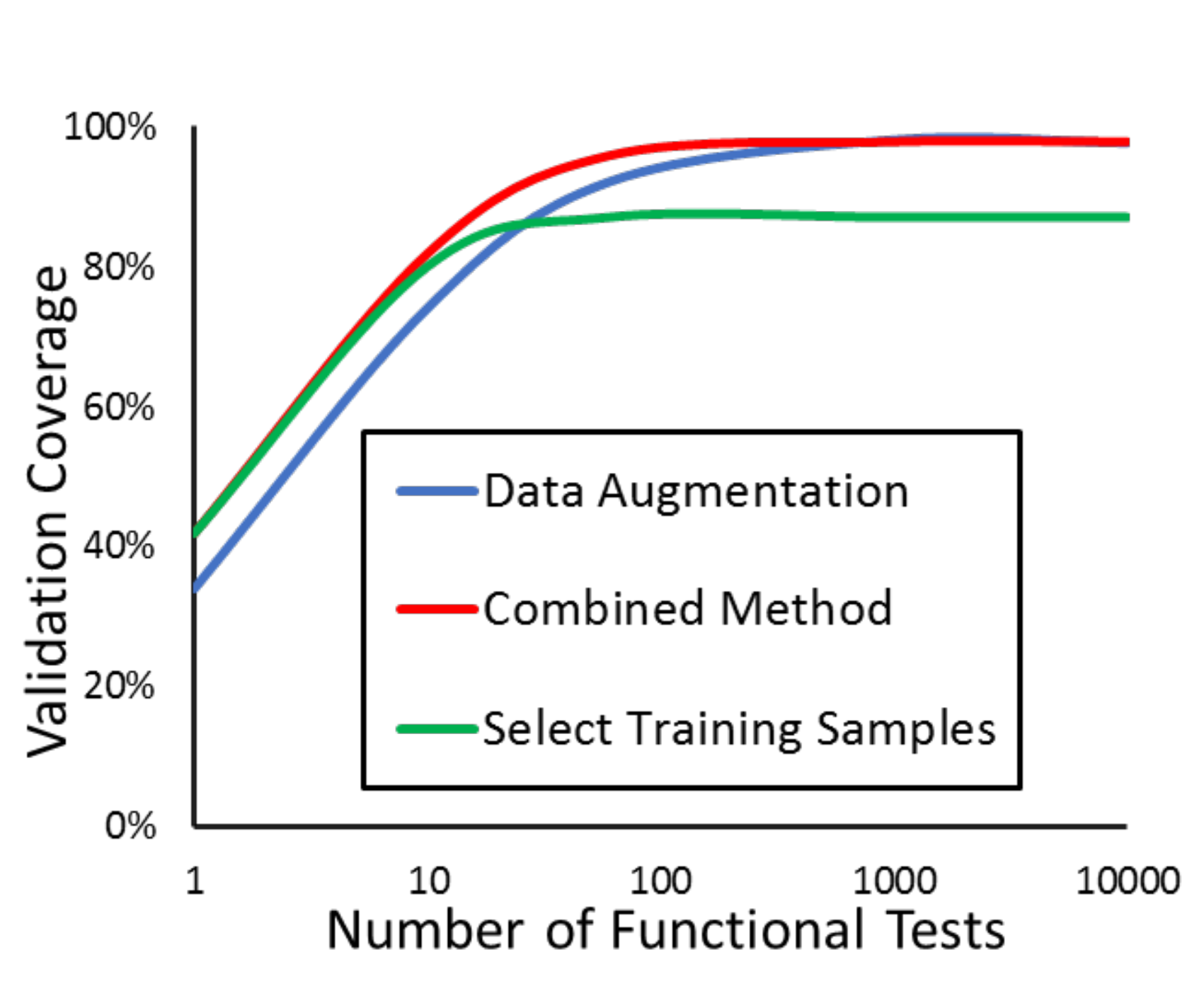}
%\captionsetup{labelfont=bf,textfont=bf}
\caption[textfont]{Validation coverage of different methods on CIFAR.}\vspace{-15pt}\label{fig:test_coverage2} %Firstly, neural vendor will generate test inputs and correct outputs, sharing with IP users. Then IP users do functionality validation by comparing Neural IP results with the shared outputs.}
%The $x$ scale is the number of training samples selected, the $y$ scale is the total detection coverage of the selected samples.}
\label{fig:test_coverage2}
\end{figure}

%The green line is the performance of algorithm 1 which selects training samples as functional tests. We can see that this method achieves a high validation coverage with just small number of selected training samples, the validation coverage of only 20 functional tests is about 84\%.

%However, the validation coverage will increase extremely slow in the late stage, the validation coverage only increases 3\% when the number of functional tests increases from 20 to 10000. Moreover, we find that there are about 8\% of parameters un-activated when using the whole training set. We analyze this phenomenon that deep neural networks are highly generalized models and some parameters or neurons are reserved for samples unseen in the training set. % some parameters are difficult to activate cause deep neural networks have redundancy and is highly generalized.

For gradient based functional test generation, the validation coverage keeps increasing until it achieves almost 100\%. This is because it can iteratively activate the un-activated parameters of DNNs by generating synthetic training samples for the remaining networks. However, it is not as efficient as selecting from training samples in the early stage, as training samples can activate more parameters than the synthetic ones. According to Fig.~\ref{fig:test_coverage2}, 10 functional tests from the training set can activate about 78\% parameters, while 10 tests generated based on gradient descent method can only activate about 66\%.

Therefore, selecting tests from the training set is efficient at the early iterations, while gradient based method is efficient in the late stages. This justifies the necessity of our combined method which takes the advantages of both methods. From the red line in Fig.~\ref{fig:test_coverage2}, we can see that our combined method achieves the best validation coverage and cost tradeoff, where 30 tests can activate 92\% parameters, while 30 training samples or synthetic samples can just activate 84\% or 76\%, respectively.

\begin{figure}[!h]
    \centering
    \includegraphics[width=0.9\columnwidth]{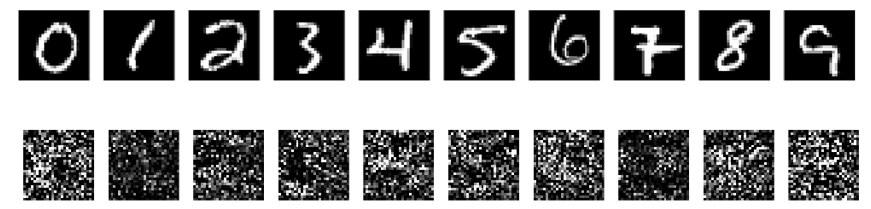}
%\captionsetup{labelfont=bf,textfont=bf}
\caption[textfont]{training samples vs. synthetic samples of MNIST.} %Firstly, neural vendor will generate test inputs and correct outputs, sharing with IP users. Then IP users do functionality validation by comparing Neural IP results with the shared outputs.}
%The $x$ scale is the number of training samples selected, the $y$ scale is the total detection coverage of the selected samples.}
\label{fig:synthetic_mnist}
\end{figure}

Moreover, to analyze the effectiveness of synthetic training samples for activating parameters, we show the real and synthetic training samples in Fig.~\ref{fig:synthetic_mnist}.
We can see that the generated samples do share some common features with the training samples of the same category. For example, the generated digit 0 in the second row has a circle in the image which is an important feature for recognizing 0. Thus, we can conclude that our gradient-based functional test generation method can efficiently generate samples containing important features for recognition and activate parameters effectively as training samples do. %thus can efficient activate neuron parameters in a neural network as the training samples do.%And for the samples generated to be highly classified by the CIFAR-10 model, we can see that each of them have different color distributions.

\begin{table*}[!htb]
\small
\begin{center}
\scalebox{1}[1]{
\begin{tabular}{|c|c|c|c|c|c|c|}
 \hline
 %\multirow{ 2}{*}{}&\multicolumn{12}{|c|}{MNIST-Tanh}\\
 %\cline{2-13}
&\multicolumn{3}{c|}{\textbf{Tests with neuron coverage}} &\multicolumn{3}{c|}{\textbf{Proposed with parameter coverage}}\\
\hline
Number of Tests&SBA& GDA&Random&SBA& GDA&Random\\
 \hline
 N=10   & 59.0\% &67.2\% &58.7\%  & 87.2\% &89.4\% & 86.3\% \\
 \hline
 N=20    &67.4\% &76.5\% &65.9\%  &91.1\% &92.5\% & 90.4\% \\
 \hline
 N=30    &76.3\% &84.1\% & 74.8\%  &93.5\% &94.7\%&92.2\%  \\
 \hline
 N=40    &82.5\% &90.2\%&80.2\%  &95.2\% &96.3\% &93.6\%  \\
 \hline
 N=50    &89.1\% &92.6\%&84.3\% &97.3\% &98.1\% &96.1\%    \\
 \hline
\end{tabular}
}
\caption{Detection rate under different perturbations on MNIST.}\label{tb:detection_mnist} %$VC$ refers the validation coverage. $DR-SBA$ and $DR-GDA$ refers the detection rate under single bias attack and gradient descent attack respectively. $DR-Random$ refers the detection rate under random perturbations.}
\end{center}
\end{table*}

\begin{table*}[!htb]
\small
\begin{center}
\scalebox{1}[1]{
\begin{tabular}{|c|c|c|c|c|c|c|}
 \hline
 &\multicolumn{3}{c|}{\textbf{Tests with neuron coverage}} &\multicolumn{3}{c|}{\textbf{Proposed with parameter coverage}} \\
\hline
Number of Tests & SBA & GDA & Random & SBA & GDA & Random\\
 \hline
 N=10    & 42.2\% &53.1\% &40.3\% & 81.0\% &82.1\% & 79.6\%\\
 \hline
 N=20    &58.3\% &67.2\% &57.6\%  &87.2\% &89.0\% & 86.2\%  \\
 \hline
 N=30    &69.2\% &76.5\% & 68.8\% &92.2\% &93.9\%&90.8\%   \\
 \hline
 N=40    &76.7\% &84.8\%&76.0\% &94.5\% &96.2\% &93.2\%  \\
 \hline
 N=50    &82.8\% &90.7\%&82.6\% &95.7\% &97.3\% &95.2\%  \\
 \hline
%\multicolumn{9}{c}{} \\
 %\multicolumn{9}{c}{(b) CIFAR-10}\\

\end{tabular}
}
\caption{Detection rate under different perturbations on CIFAR.}\label{tb:detection_cifar}\vspace{-10pt} %$VC$ refers the validation coverage. $DR-SBA$ and $DR-GDA$ refers the detection rate under single bias attack and gradient descent attack respectively. $DR-Random$ refers the detection rate under random perturbations.}
\end{center}
\end{table*}

\subsection{Perturbation Detection Rate}
In this section, we evaluate the performance of the proposed validation scheme considering its detection rate under malicious and random parameter perturbations. The malicious perturbations are generated according to the attacks proposed in \cite{liu2017fault} and the random perturbations are to add gaussian noises. We implement each kind of parameter perturbation 10000 times against the MNIST and CIFAR-10 models, and then calculate the detection rate by observing whether the perturbations will change the DNN outputs of the generated functional tests.
In order to justify the necessity of considering parameter coverage instead of neuron coverage, we compare our combined functional test generation method with the hardware testing technique that only considers neuron coverage \cite{ma2018combinatorial}. It should be noted that hardware testing cannot be used in this case as users have no access to intermediate DNN results.

Table~\ref{tb:detection_mnist} and~\ref{tb:detection_cifar} show the detection rates for MNIST and CIFAR-10 under single bias attack (SBA), gradient descent attack (GDA)~\cite{liu2017fault} and random perturbations, respectively.
%The results are in Table~\ref{fig:detection}. $DR-SBA$ and $DR-GDA$ refer to the detection rate of malicious perturbations generated by single bias attack and gradient descent attack respectively in \cite{liu2017fault}. $DR-Random$ refers to the detection rate of random perturbations. $VC$ is the validation coverage defined in equation , the percentage of activated parameters.
%\subsubsection{Detection Rate under Malicious Perturbation}
%For the malicious perturbation proposed in \cite{liu2017fault}, there are two kinds of attack methods. One is single bias attack (SBA) that only perturbs one bias parameter to misclassify an input pattern, but the perturbation added to the bias should be large enough. The other is the gradient descent attack (GDA) that perturbs many parameters with small values. We evaluate the detection rate of the validation scheme under these two malicious attack methods. The results are shown in Table 2, namely $DR-SBA$ and $DR-GDA$.
We can see that our combined test generation method achieves 87.2\% and 89.0\% detection rates under SBA and GDA respectively with only 20 functional tests for the CIFAR-10 model. Comparing with the test generation method considering neuron coverage, it performs worse than our combined method, achieving much lower detection rate with the same number of functional tests. Even though all neurons are covered by test cases, it is not necessarily to cover all parameters. This justifies the necessity of considering parameter coverage in our proposed solution.

\section{Conclusions}\label{sec:conclusion}
In this paper, we propose a practical validation scheme for DNN IPs without showing users model parameters. The idea is to generate a small number of functional tests to largely activate model parameters. Then the perturbations on them will propagate to the outputs and be detected. Considering the large amounts of parameters and highly non-linearity of DNNs, it is very challenging to solve this problem. In this work, we first propose to judiciously select test cases from the training set and when this method becomes inefficient, we present a gradient-based new test generation techniques. Finally, these two methods are combined in a unified way to achieve both advantages. Experimental results show that our solution achieves a good trade off between validation coverage and cost, and can effectively detect malicious and random perturbations with a reasonable number of tests.% a high detection coverage with just small number of functional tests. Then we evaluate the performance of the proposed validation scheme considering the detection rate under malicious perturbations and random perturbations. The experimental results show that our scheme can detect almost all the perturbations in the MNIST and CIFAR-10 model with a low validation cost.

\section*{Acknowledgement}
\label{acknowledgement}

This work was supported in part by the General Research Fund (GRF) of Hong Kong Research Grants Council (RGC) under Grant No. 14205018 and in part by National Natural Science Foundation of China under Grant No. 61432017 and No. 61532017.

\bibliographystyle{ieeetr}
\bibliography{ref}

\end{document}